\documentclass{article}

\usepackage{arxiv}

\usepackage[utf8]{inputenc} 
\usepackage[T1]{fontenc}    
\usepackage{hyperref}       
\usepackage{url}            
\usepackage{booktabs}       
\usepackage{amsfonts}       
\usepackage{nicefrac}       
\usepackage{microtype}      
\usepackage{lipsum}

\usepackage{amsmath}
\usepackage{graphicx}

\title{Interpretable Vision Transformers in Monocular Depth Estimation via SVDA}

\author{
  Vasileios Arampatzakis\\
  Dept. Electrical and Computer Engineering\\
  Democritus University of Thrace\\
  Athena Research Center\\
  University Campus at Kimmeria\\
  67100 Xanthi, Greece\\
  \texttt{vaarampa@ee.duth.gr} \\
   \And
 George Pavlidis\\
Athena Research Center\\
University Campus at Kimmeria\\
67100 Xanthi, Greece\\
  \texttt{gpavlid@athenarc.gr} \\
  \And
Nikolaos Mitianoudis\\
Dept. Electrical and Computer Engineering\\
Democritus University of Thrace\\
University Campus at Kimmeria\\
67100 Xanthi, Greece\\
  \texttt{nmitiano@ee.duth.gr} \\
    \And
Nikos Papamarkos\\
Dept. Electrical and Computer Engineering\\
Democritus University of Thrace\\
University Campus at Kimmeria\\
67100 Xanthi, Greece\\
  \texttt{papamark@ee.duth.gr} \\
}

\begin{document}
\maketitle

\begin{abstract}
Monocular depth estimation is a central problem in computer vision with applications in robotics, AR, and autonomous driving, yet the self-attention mechanisms that drive modern Transformer architectures remain opaque. We introduce SVD-Inspired Attention (SVDA) into the Dense Prediction Transformer (DPT), providing the first spectrally structured formulation of attention for dense prediction tasks. SVDA decouples directional alignment from spectral modulation by embedding a learnable diagonal matrix into normalized query–key interactions, enabling attention maps that are intrinsically interpretable rather than post-hoc approximations. Experiments on KITTI and NYU-v2 show that SVDA preserves or slightly improves predictive accuracy while adding only minor computational overhead. More importantly, SVDA unlocks six spectral indicators that quantify entropy, rank, sparsity, alignment, selectivity, and robustness. These reveal consistent cross-dataset and depth-wise patterns in how attention organizes during training, insights that remain inaccessible in standard Transformers. By shifting the role of attention from opaque mechanism to quantifiable descriptor, SVDA redefines interpretability in monocular depth estimation and opens a principled avenue toward transparent dense prediction models.
\end{abstract}


\section{Introduction}
\label{sec:intro}

Monocular depth estimation, the task of predicting per-pixel depth from a single RGB image, is an inherently ill-posed problem with wide-ranging applications in autonomous navigation, augmented reality, and 3D scene reconstruction \cite{arampatzakis2024review}. Recent advances in vision Transformers, such as the Dense Prediction Transformer (DPT), have demonstrated strong performance in this domain by leveraging global self-attention and multi-scale feature fusion \cite{ranftl2021vision}. However, despite their empirical success, these models remain largely opaque, with attention mechanisms that are difficult to interpret and lacking in geometric or semantic structure \cite{dosovitskiy2020image, wiegreffe2019attention, chefer2021transformer, raghu2021vision}.

While dense prediction tasks such as depth estimation require fine-grained spatial reasoning, current Transformer-based models often yield attention maps that are dense and uninterpretable. This opacity limits our ability to understand the model's internal representations and compromises trust in high-stakes applications such as robotics, autonomous driving, and scientific measurement \cite{rudin2019stop}. Prior work in image classification has shown that standard attention maps may fail to reflect semantically meaningful interactions \cite{wiegreffe2019attention}, and similar concerns apply to dense regression tasks \cite{rudin2019stop}. \textit{Addressing this challenge calls for attention mechanisms that promote interpretability, structured behavior, and semantic alignment with scene geometry.}

In this work, we integrate the SVD-Inspired Attention (SVDA) mechanism \cite{arampatzakis2025geometry} into the DPT framework to enhance interpretability in monocular depth estimation. SVDA replaces the standard dot-product formulation with a spectral attention design, consisting of row-wise normalized query/key projections and a learnable diagonal spectral modulation matrix~$\Sigma$. Inspired by Singular Value Decomposition (SVD), this structure decouples directional attention from spectral importance---a property especially beneficial for depth prediction, where spatial continuity and semantic coherence are critical. \textit{This formulation produces attention maps that are intrinsically interpretable, requiring only minimal architectural change and incurring a small runtime overhead while preserving predictive accuracy.} In addition to enhancing attention structure, SVDA unlocks a suite of spectral indicators that expose the inner mechanics of attention layers. These indicators establish a coherent diagnostic framework for probing interpretability, directional alignment, and information flow across layers and heads.

This work explores the integration of SVDA into the DPT architecture for monocular depth estimation, with a focus on enhancing interpretability through spectral structure. We conduct experiments on two diverse datasets, KITTI and NYU-v2. We evaluate interpretability using six spectral indicators---spectral entropy, effective rank, angular alignment, selectivity index, spectral sparsity, and perturbation robustness---that quantify structure and concentration in the attention maps.

Our key contributions are:
\begin{itemize}
    \item We adapt the SVDA mechanism to the Dense Prediction Transformer (DPT), introducing spectral and geometric regularization into attention layers for monocular depth estimation.
    \item We demonstrate SVDA's diagnostic framework for analyzing attention dynamics, sparsity, and geometric alignment in monocular depth estimation.
    \item We demonstrate that SVDA-powered monocular depth estimation maintains competitive accuracy while enabling the interpretability benefits using extensive experiments on two depth estimation benchmarks.
\end{itemize}

This approach bridges spectral theory with practical vision architectures, enabling interpretable Transformer behavior in spatially continuous tasks such as depth estimation.

\section{Related Work}

Monocular Depth Estimation (MDE) has advanced rapidly with deep learning, evolving from early convolutional models \cite{laina2016deeper, fu2018deep} to more recent Transformer-based architectures. The Dense Prediction Transformer (DPT) \cite{ranftl2021vision} introduced a ViT-style encoder-decoder framework capable of modeling long-range dependencies and multi-scale semantics, setting a new benchmark in dense regression. AdaBins \cite{bhat2021adabins} extended this line by discretizing depth into adaptive bins, significantly improving resolution and consistency.

Several follow-up works apply Transformers to MDE using architectural innovations or self-supervision. MonoViT \cite{zhao2022monovit} integrates ViTs into a self-supervised framework, while Depthformer \cite{agarwal2022depthformer} combines multi-scale attention with depth-aware fusion strategies. CATNet \cite{gao2024catnet} and PCTDepth \cite{xia2024pctdepth} explore hybrid CNN–Transformer decoders to balance local detail with global context. However, these approaches largely prioritize performance and architectural efficiency rather than interpretability or structural transparency of attention.


Interpretability of attention mechanisms remains a fundamental concern in Transformer models, particularly for dense prediction tasks. While attention is often treated as an explanatory tool, foundational studies have cautioned against over-reliance on raw attention maps for interpretability \cite{wiegreffe2019attention, serrano2019attention}. In the vision domain, Chefer et al. \cite{chefer2021transformer} extend attention-based explanation with attribution propagation, and Raghu et al. \cite{raghu2021vision} analyze how attention evolves across layers. Nevertheless, these works primarily address classification tasks. Beyond architectural design, there have been attempts to introduce interpretability directly into monocular depth estimation. You et al.~\cite{you2021interpretable} proposed a unit-level approach that enforces neuron selectivity to specific depth ranges, thereby assigning semantic meaning to hidden activations. While effective in attributing function at the neuron scale, their method does not capture the global spectral organization of attention across layers. Gipiškis et al.~\cite{gipiskis2024xaiSeg} surveyed explainable AI methods in segmentation, emphasizing the use of attribution maps and visual explanation techniques to guide model design and compression. This approach remains largely prescriptive, whereas SVDA is diagnostic: it reveals how attention naturally structures itself without altering the model to conform to external heuristics. More recently, Cirillo et al.~\cite{sheddingDepth2025} assessed attribution methods in monocular depth estimation, introducing a framework for evaluating the fidelity of saliency-based explanations. Their work evaluates post-hoc input-level explanations, while SVDA offers intrinsic, layerwise interpretability through spectral indicators. In contrast to these methods, SVDA does not aim to optimize for interpretability or impose semantic roles, but to provide a principled decomposition that describes how attention spectra evolve across training and depth.


Recent works introduce geometric or structural priors into ViT architectures to improve depth estimation. Ruhkamp et al. \cite{ruhkamp2021geometry} propose spatial-temporal attention modules informed by geometric constraints to enhance consistency. TinyDepth \cite{cheng2024tinydepth} and METER \cite{papa2024meter} inject inductive biases via lightweight attention decoders, while CATNet \cite{gao2024catnet} and PCTDepth \cite{xia2024pctdepth} combine local convolutional features with global Transformer heads for hybrid attention.

Although these methods incorporate structural biases---such as local/global fusion or convolutional regularization---\textit{none apply spectral analysis or principled decomposition to inform or constrain the attention mechanism itself}. They optimize architectural design but do not address the interpretability or internal organization of attention maps.


Spectral decompositions, such as low-rank or kernel approximations for efficient inference \cite{choromanski2021rethinking}, have been explored primarily for scalability but have rarely been applied to enhance interpretability in vision models. To the best of our knowledge, only a recent study introducing SVDA \cite{arampatzakis2025geometry} enables direct control over directional and spectral contributions within attention matrices.

In this work, we extend SVDA to the domain of monocular depth estimation. Unlike prior MDE models that rely on architectural heuristics or fusion strategies, SVDA embeds spectral structure directly into the attention mechanism, enabling interpretability. To our knowledge, this is the first application of spectral analysis to systematically structure and interpret attention mechanisms in monocular depth estimation.

\section{SVDA in Monocular Depth Estimation}

SVDA imposes a structured decomposition on the attention mechanism by separating directional encoding from spectral modulation. Drawing inspiration from the Singular Value Decomposition (SVD), it modifies conventional dot-product attention by applying $\ell_2$-normalized query and key projections together with a learnable, diagonal spectral matrix~$\Sigma$ unique to each attention head. The core SVDA attention is defined as follows. Let \( Q, K \in \mathbb{R}^{n \times d_k} \) be row-wise \( \ell_2 \)-normalized projections (i.e., each row \( Q_i, K_j \) satisfies \( \|Q_i\|_2 = \|K_j\|_2 = 1 \)), and let \( \Sigma \in \mathbb{R}^{d_k \times d_k} \) be a learned diagonal matrix. Define the attention matrix as:
\begin{equation}
\label{eq:attention}
A = \mathrm{softmax}\left( \frac{Q \Sigma K^\top}{\sqrt{d_k}} \right)
\end{equation}
where $Q, K \in \mathbb{R}^{n \times d_k}$ denote the query and key matrices, normalized row-wise, and $\Sigma \in \mathbb{R}^{d_k \times d_k}$ is a trainable diagonal matrix that adjusts the contribution of each latent dimension. This structure mirrors the SVD-inspired decomposition of attention into directional and spectral terms, facilitating interpretable and structured attention maps, subject to subsequent softmax normalization and head-wise processing.

Our implementation closely follows the embedding pipeline of the original DPT architecture~\cite{ranftl2021vision} for direct comparison. The image is first split into non-overlapping $P \times P$ patches, flattened into vectors, and passed through a learned linear projection to generate patch embeddings. Fixed-length learnable positional embeddings are added to patch tokens to maintain spatial order. These token sequences are then processed by a ViT encoder. The only departure from the baseline DPT model is the replacement of the self-attention mechanism with SVDA, as described in Eq.\ref{eq:attention} and detailed in~\cite{arampatzakis2025geometry}.

As defined in the original publication of SVDA, the interpretability indicators to be also used here include

\begin{itemize}
\item Spectral Entropy \( \mathcal{H}_\Sigma = -\sum p_i \log p_i \): Information indicator that captures focus/disorder in latent dimension usage

\item Effective Rank \( \mathrm{rank}_{\text{eff}}(\Sigma) = e^{\mathcal{H}_\Sigma} \): Scalar Indicator that measures compressibility or dimension reduction capacity

\item Angular Alignment \( \cos \theta_{ij} = Q_i \cdot K_j \): Pseudo-metric that reflects semantic closeness in latent space

\item Selectivity Index \( \mathcal{S}(A_i) = 1 - \frac{\sum_j A_{ij}^2}{\left(\sum_j A_{ij}\right)^2} \): Statistical indicator that reflects the token-level sparsity and focus in attention matrix

\item Spectral Sparsity \( \mathcal{P}(\Sigma) = \frac{|\{ \sigma_i : |\sigma_i| < \varepsilon \}|}{d_k} \): Structural indicator that indicates inactive latent directions and pruning potential

\item Perturbation Robustness \( \Delta A = \|A(x) - A(x + \delta)\|_F \): Metric that quantifies behavioral consistency under small perturbations 
\end{itemize}

\section{Results and Structural Indicators}

Beyond accuracy and runtime, the evaluation examines interpretability and structural organization of attention via the SVDA indicators. Those indicators are computed for each head and layer at every epoch, enabling both temporal tracking and depth-wise analysis of attention dynamics. 

\begin{figure}[!t]
\centering
\begin{minipage}{0.48\linewidth}\centering \textbf{KITTI}\\[0.25em]
\includegraphics[width=\linewidth]{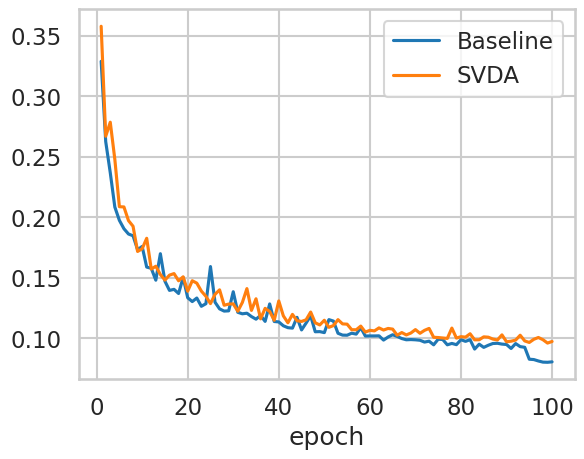}
\end{minipage}\hfill
\begin{minipage}{0.48\linewidth}\centering \textbf{NYU-v2}\\[0.25em]
\includegraphics[width=\linewidth]{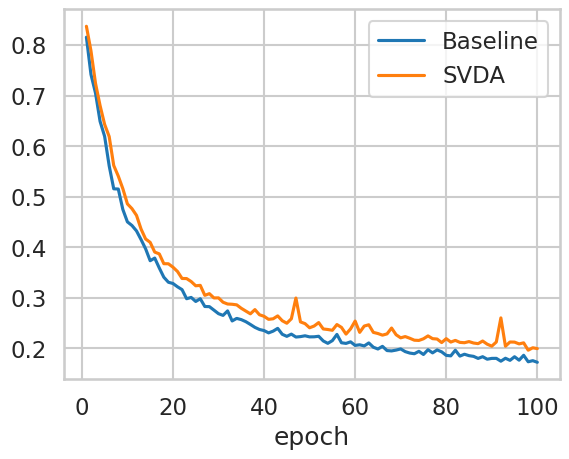}
\end{minipage}
\caption{Validation loss over epochs for baseline and SVDA models across datasets. The models show nearly identical learning trajectories.}
\label{fig:mde_svda_loss}
\end{figure}

Consistent convergence behavior is observed across KITTI and NYU‑v2. As illustrated in \figurename~\ref{fig:mde_svda_loss}, validation losses are nearly identical between SVDA‑enhanced and the baseline model, indicating stable optimization and no signs of overfitting. In addition, \tablename~\ref{tab:table-evaluation-svda-kitti} and \tablename~\ref{tab:table-evaluation-svda-nyuv2} summarize the benchmark comparison results.

\begin{table}[!t]
\centering
\caption{Benchmark comparison on KITTI, evaluated at the checkpoint selected by validation AbsRel (lower is better). Selected epochs: Baseline = 65, SVDA = 66.}
\setlength{\tabcolsep}{3pt} 
\begin{tabular}{lccccccc}
\toprule
{\textbf{Method}} & \textbf{AbsRel} & \textbf{SqRel} & \textbf{RMSE} & \textbf{RMSE\textsubscript{log}} & \textbf{sRMSE\textsubscript{log}} & $\boldsymbol\delta_1$ \\
\midrule
Baseline & 0.058 & \textbf{0.001} & \textbf{0.010} & 0.037 & \textbf{0.002} & 0.976 \\ 
\midrule
\textbf{SVDA (ours)}& \textbf{0.056} & \textbf{0.001} & 0.011 & \textbf{0.035} & \textbf{0.002} & \textbf{0.979} \\ 
\bottomrule
\end{tabular}
\label{tab:table-evaluation-svda-kitti}
\end{table}

\begin{table}[!t]
\centering
\caption{Benchmark comparison on NYU-v2, evaluated at the checkpoint selected by validation AbsRel (lower is better). Selected epochs: Baseline = 93, SVDA = 95.}
\setlength{\tabcolsep}{3pt} 
\begin{tabular}{lccccccc}
\toprule
{\textbf{Method}} & \textbf{AbsRel} & \textbf{SqRel} & \textbf{RMSE} & \textbf{RMSE\textsubscript{log}} & \textbf{sRMSE\textsubscript{log}} & $\boldsymbol\delta_1$ \\
\midrule
Baseline & 0.133 & \textbf{0.010} & 0.079 & \textbf{0.093} & \textbf{0.013} & 0.865 \\ 
\midrule
\textbf{SVDA (ours)}& \textbf{0.124} & \textbf{0.010} & \textbf{0.071} & 0.109 & \textbf{0.013} & \textbf{0.872} \\ 
\bottomrule
\end{tabular}
\label{tab:table-evaluation-svda-nyuv2}
\end{table}

SVDA introduces only $0.01\%$ additional parameters compared to the baseline, but incurs a runtime increase of approximately $15.79\%$. This overhead is due to additional $\ell_2$-normalization and spectral modulation steps. This overhead is acceptable given the benefits of SVDA, structurally organized attention with explicit spectral regularization and improved interpretability-while leaving the remainder of the DPT-style pipeline unchanged. Notably, in our experiments, SVDA reported lower MAC counts than the baseline ($169.67$~GMac vs.\ $182.05$~GMac), a $6.80\%$ reduction, indicating that the overhead is largely due to implementation inefficiencies rather than fundamental complexity. Given the interpretability benefits and stable performance, this trade‑off is acceptable for applications where transparency is important~\cite{dao2022flashattention}.

\begin{figure*}[!t]
\centering

\setlength{\tabcolsep}{2pt}

\begin{tabular}{p{0.22\textwidth} p{0.22\textwidth} p{0.22\textwidth} p{0.22\textwidth}}

\multicolumn{2}{c}{\textbf{KITTI}} & \multicolumn{2}{c}{\textbf{NYU-v2}} \\

{\centering \includegraphics[width=\linewidth]{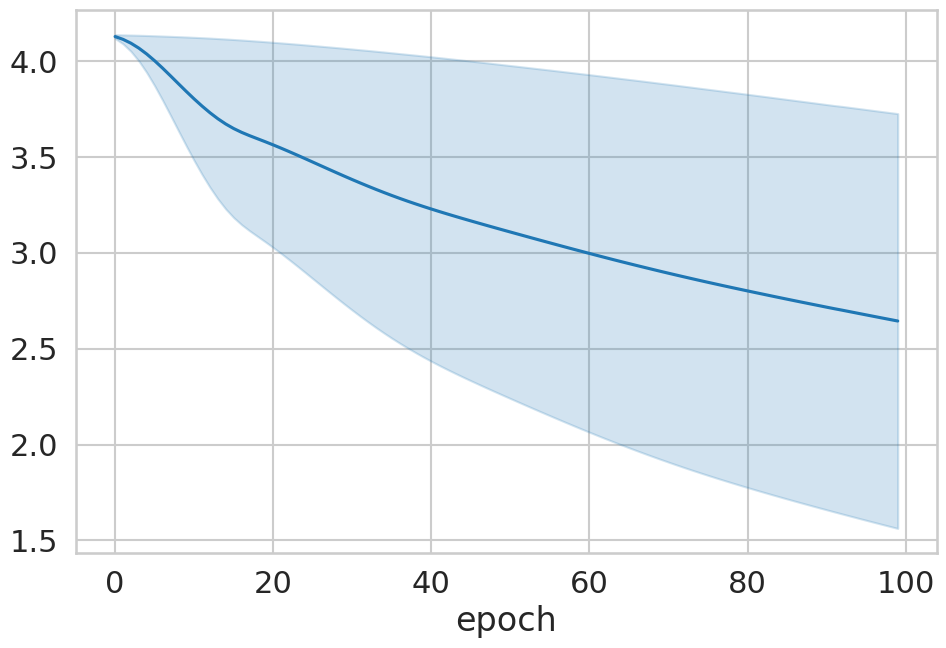}\par} &
{\centering \includegraphics[width=\linewidth]{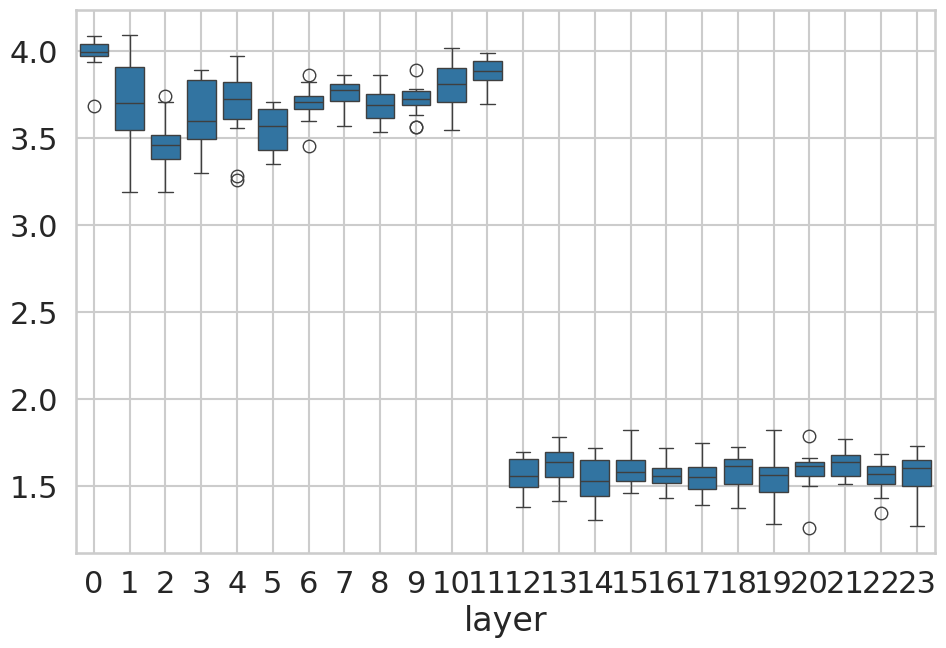}\par} &
{\centering \includegraphics[width=\linewidth]{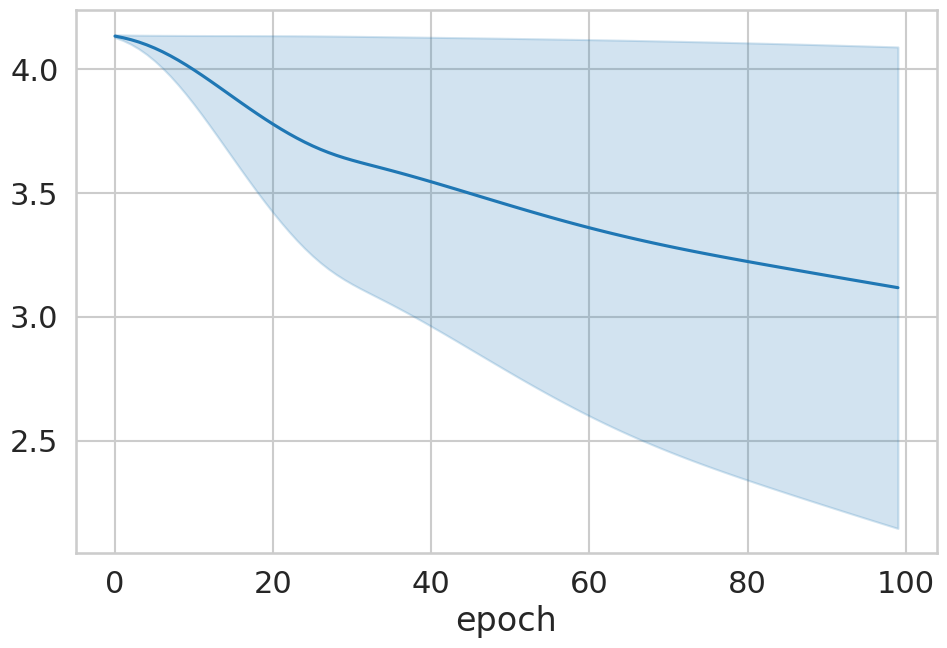}\par} &
{\centering \includegraphics[width=\linewidth]{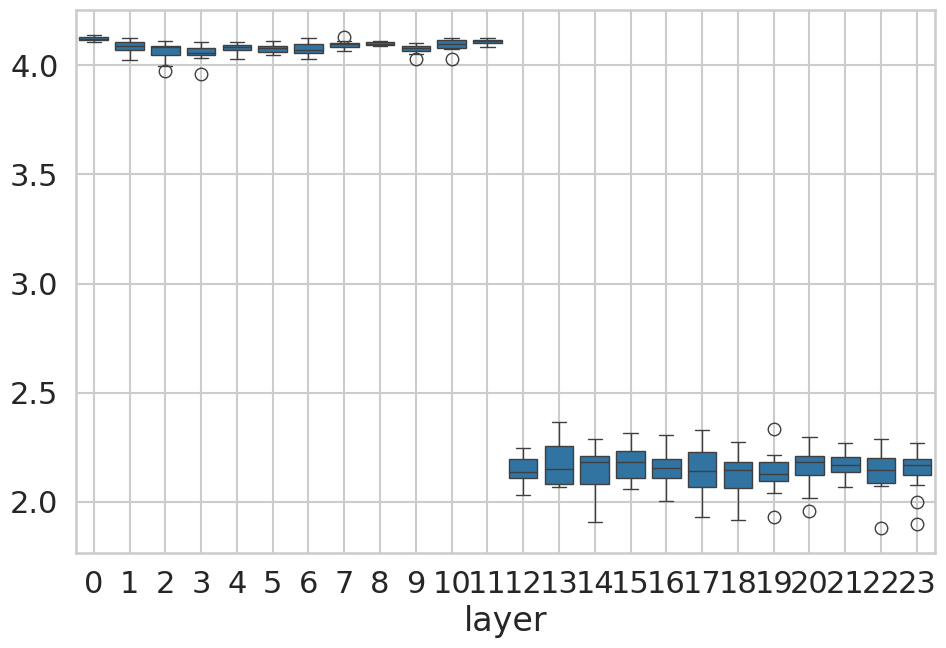}\par} \\[0.0 ex]

\multicolumn{4}{c}{\centering{\textbf{(a)}}} \\[1.0ex]

{\centering \includegraphics[width=\linewidth]{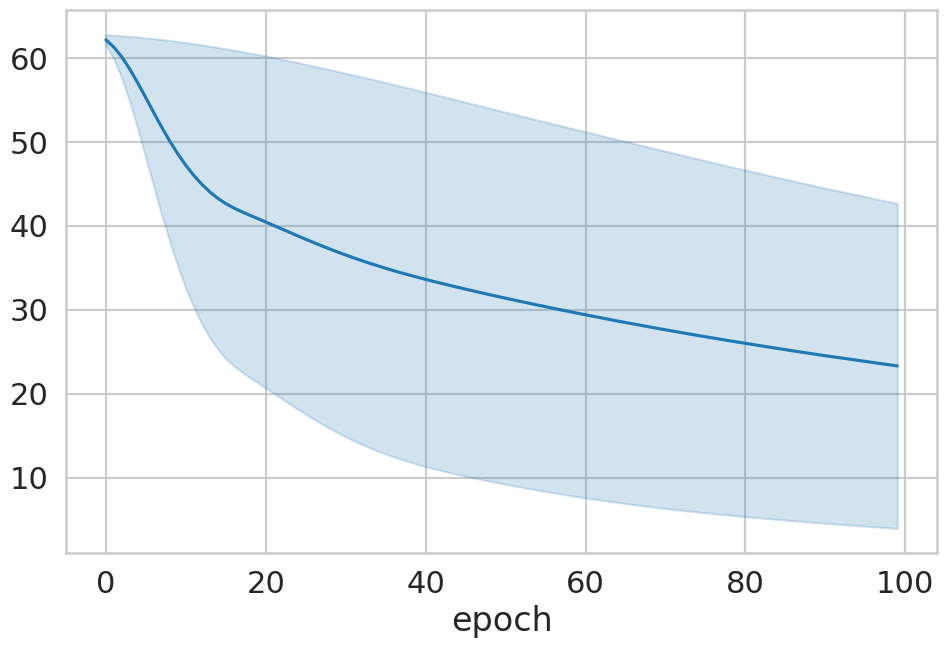}\par} &
{\centering \includegraphics[width=\linewidth]{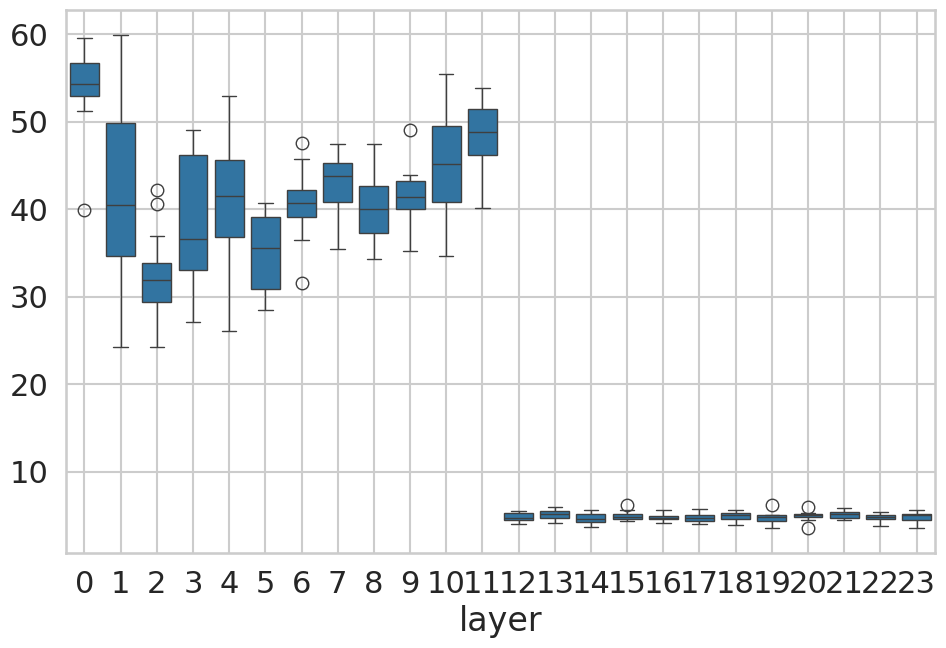}\par} &
{\centering \includegraphics[width=\linewidth]{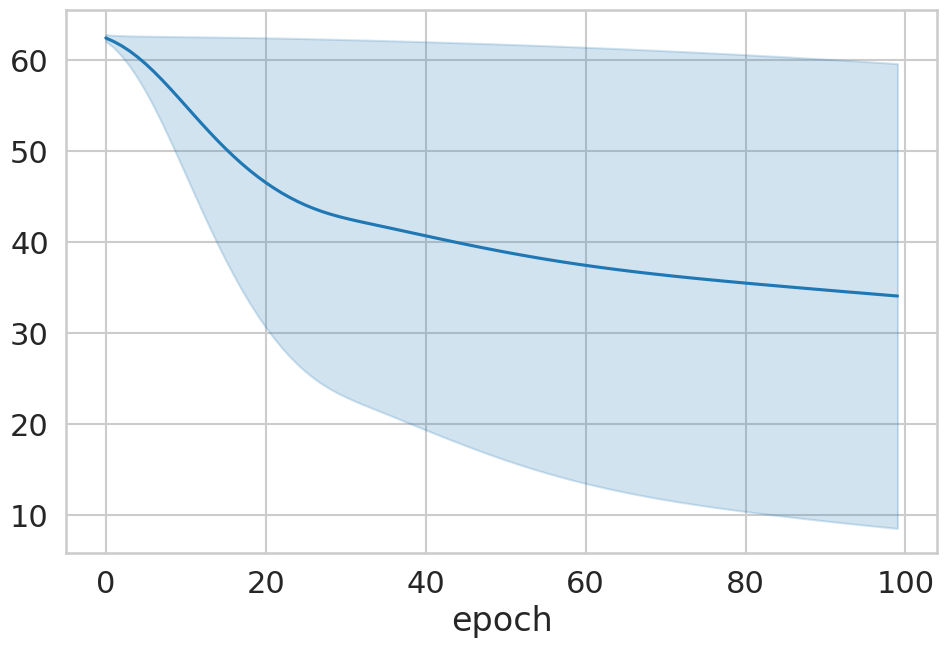}\par} &
{\centering \includegraphics[width=\linewidth]{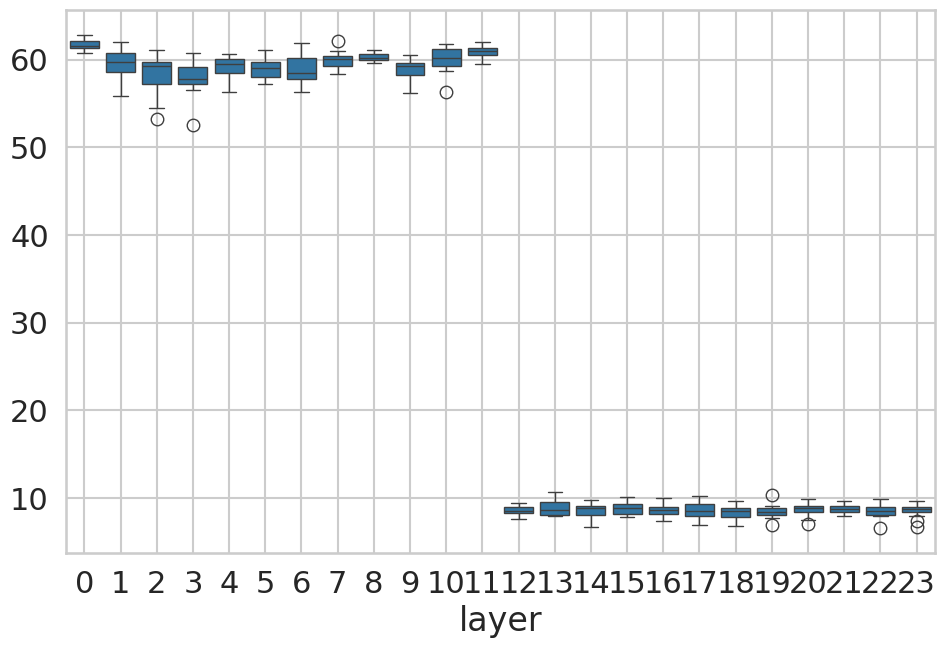}\par} \\[0.0 ex]

\multicolumn{4}{c}{\textbf{(b)}} \\[1.0ex]

{\centering \includegraphics[width=\linewidth]{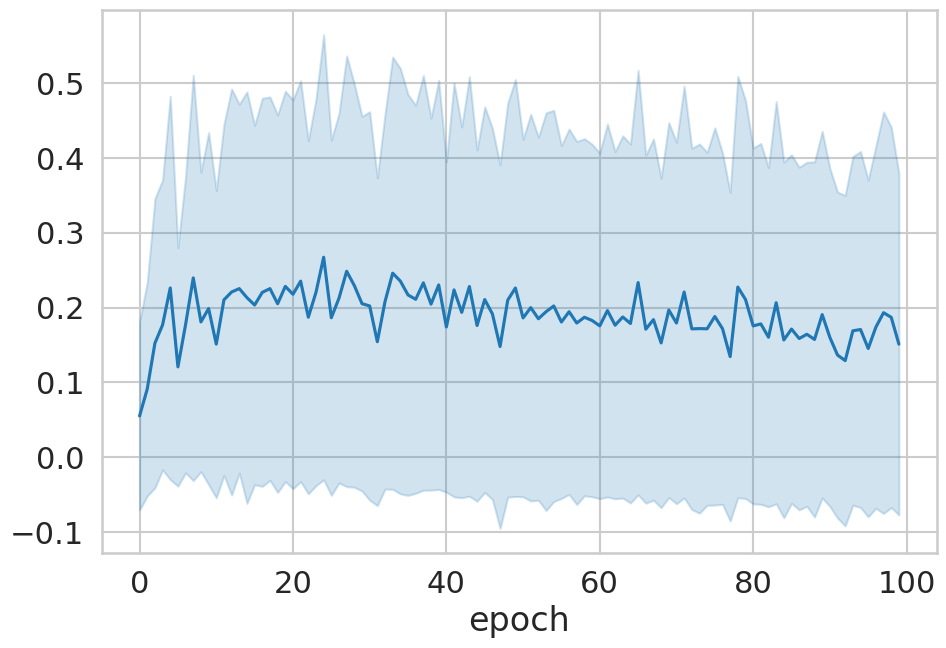}\par} &
{\centering \includegraphics[width=\linewidth]{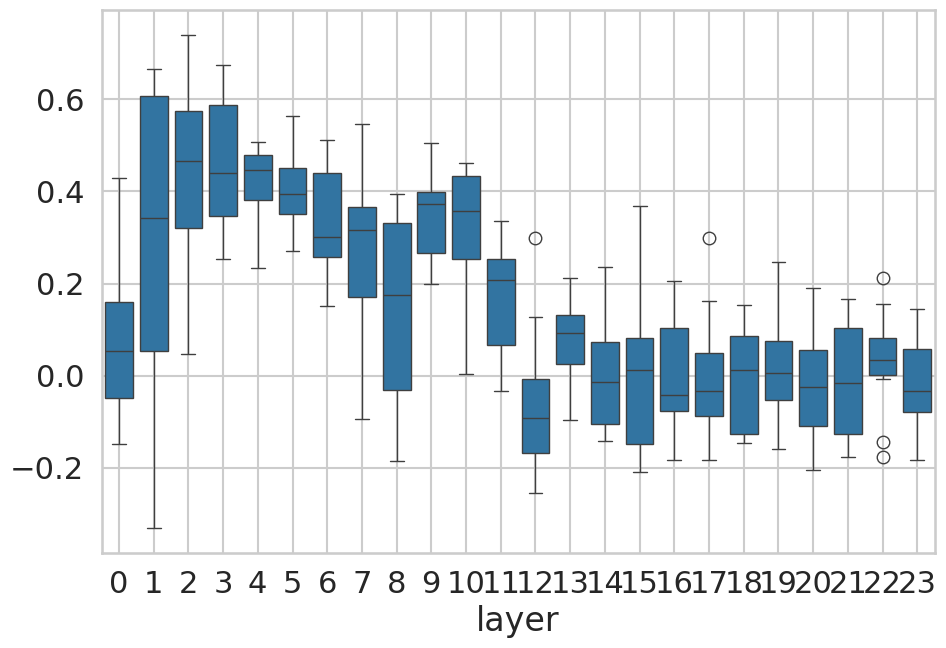}\par} &
{\centering \includegraphics[width=\linewidth]{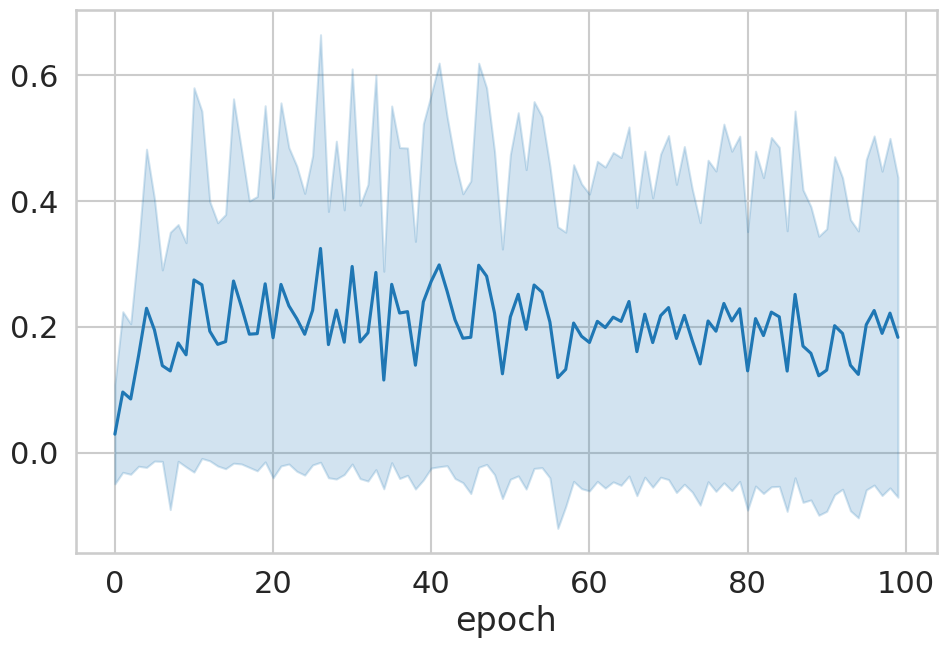}\par} &
{\centering \includegraphics[width=\linewidth]{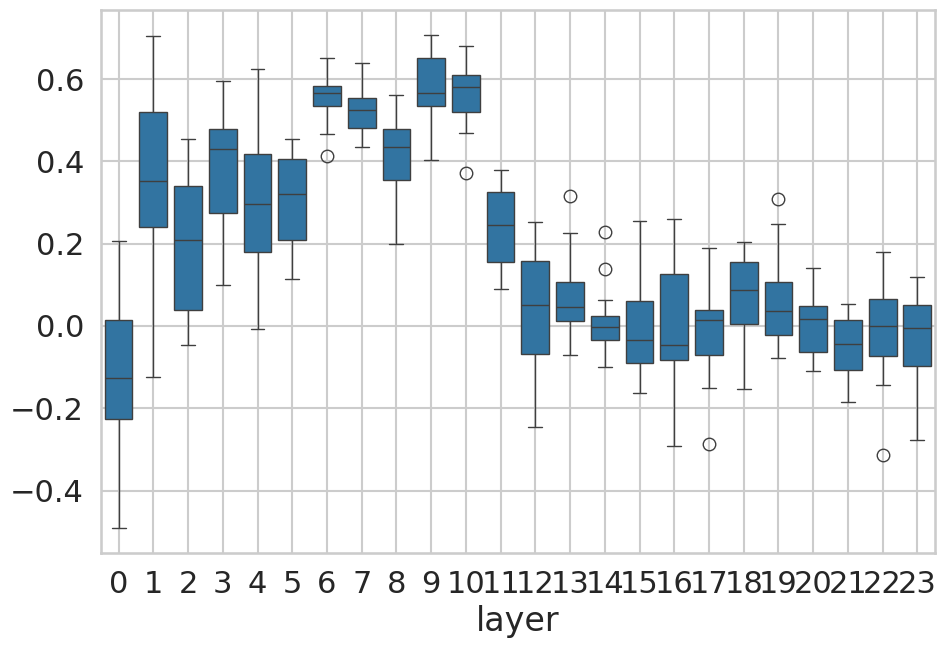}\par} \\[0.0 ex]

\multicolumn{4}{c}{\textbf{(c)}} \\[1.0ex]

{\centering \includegraphics[width=\linewidth]{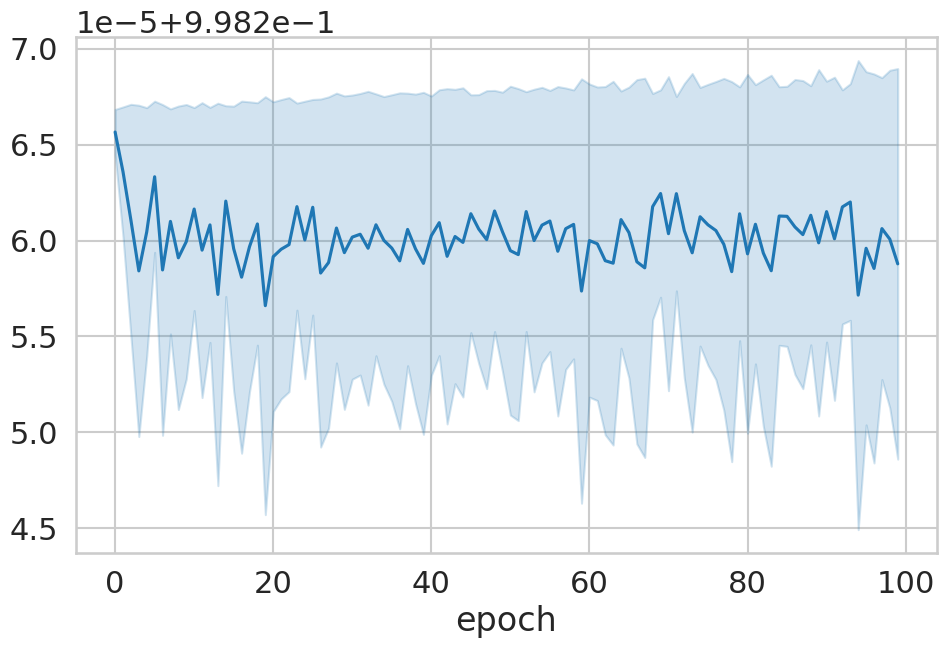}\par} &
{\centering \includegraphics[width=\linewidth]{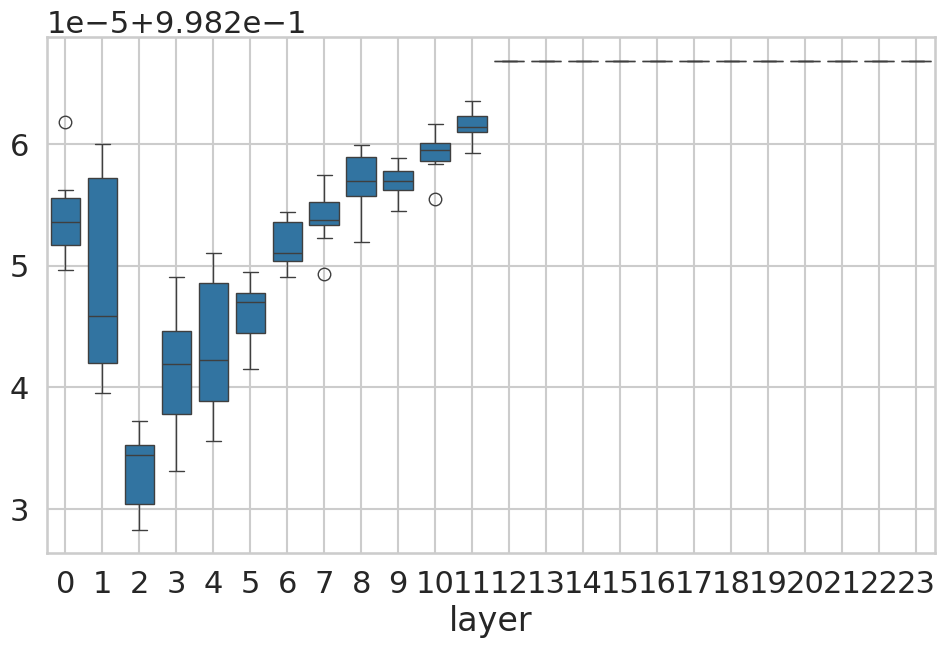}\par} &
{\centering \includegraphics[width=\linewidth]{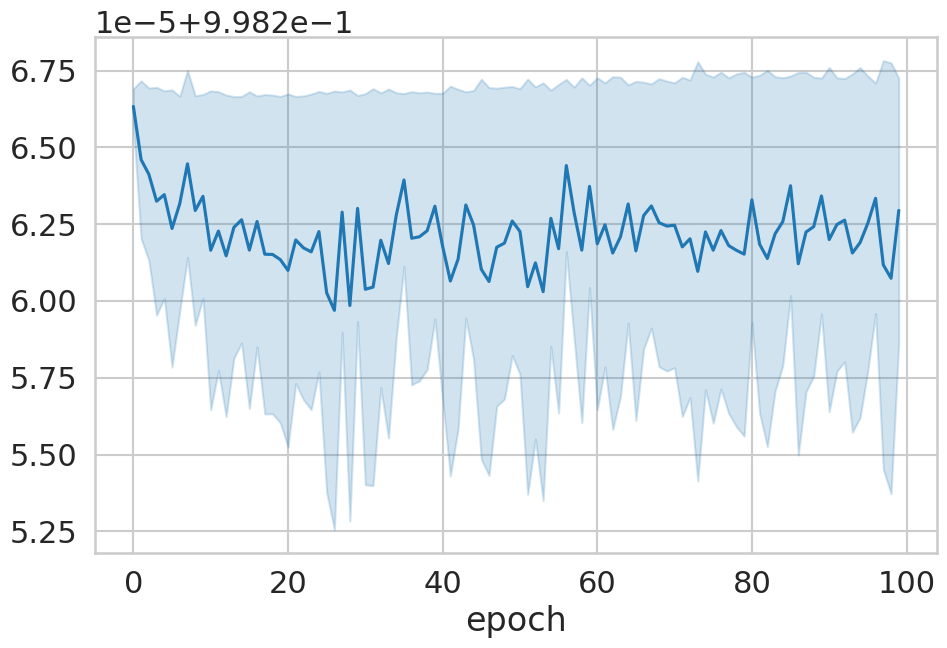}\par} &
{\centering \includegraphics[width=\linewidth]{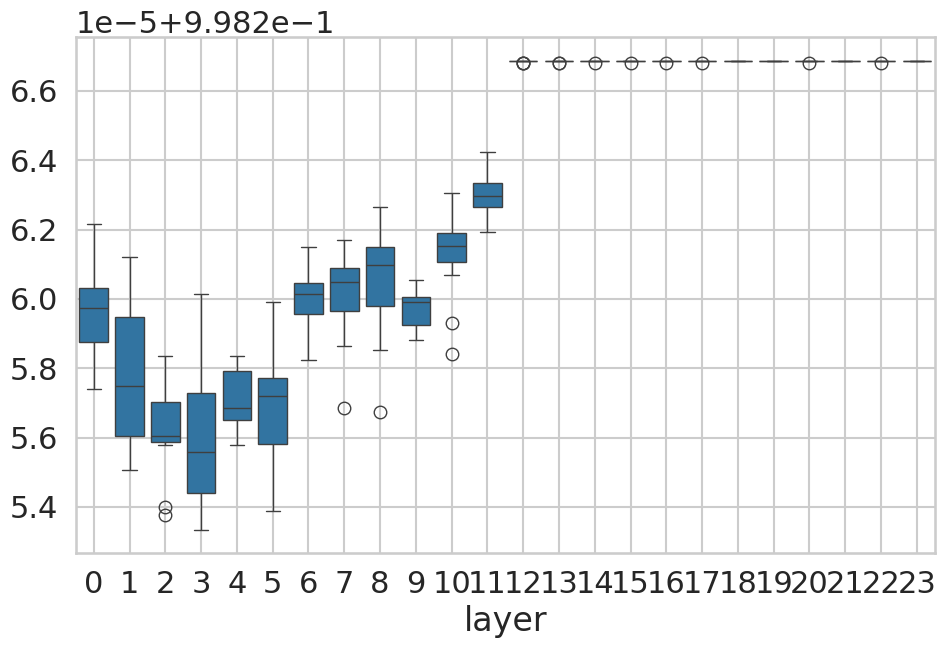}\par} \\[0.0 ex]

\multicolumn{4}{c}{\textbf{(d)}} \\[1.0ex]

{\centering \includegraphics[width=\linewidth]{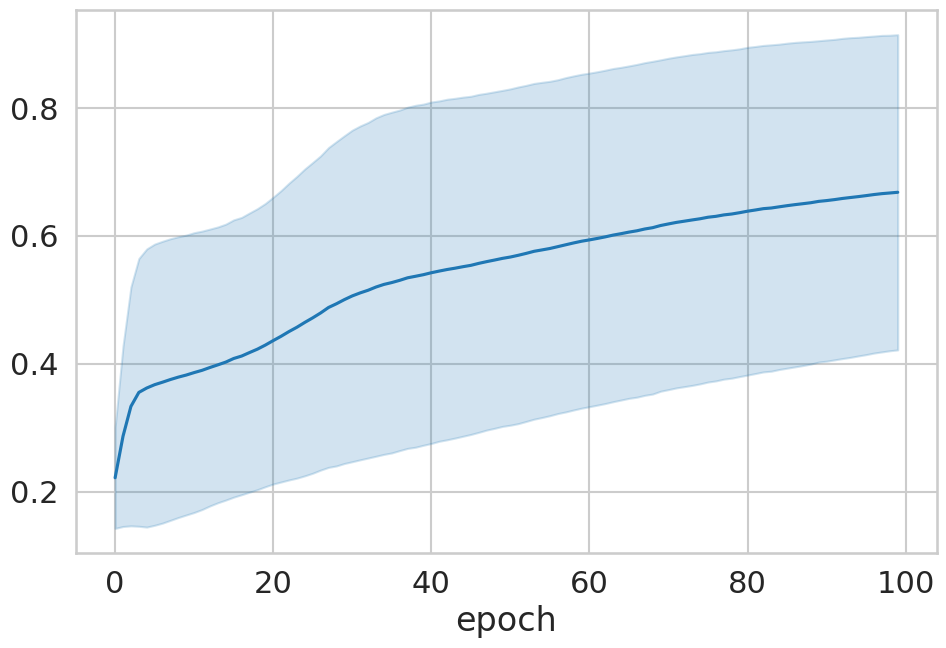}\par} &
{\centering \includegraphics[width=\linewidth]{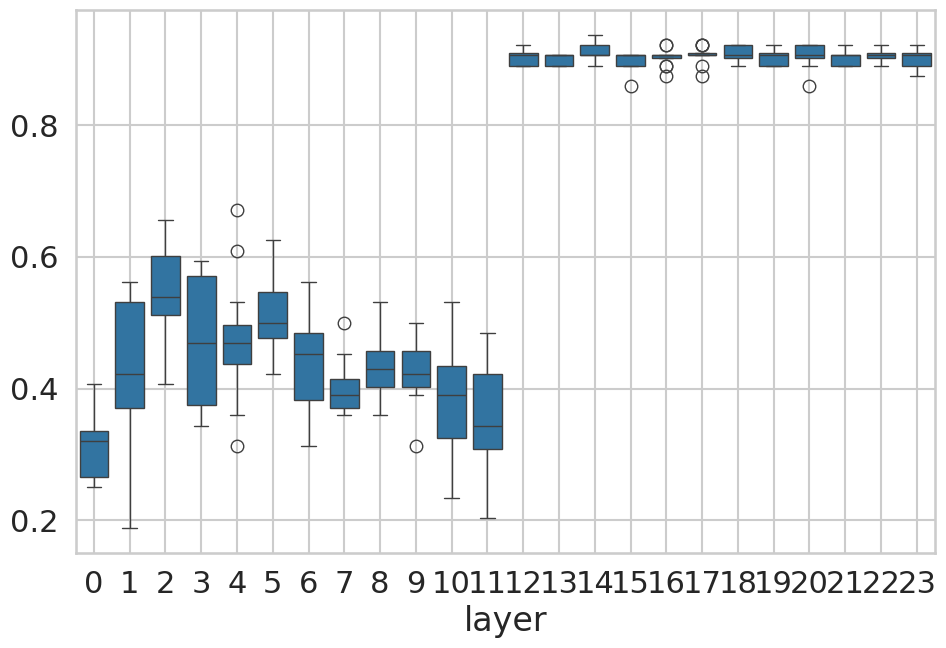}\par} &
{\centering \includegraphics[width=\linewidth]{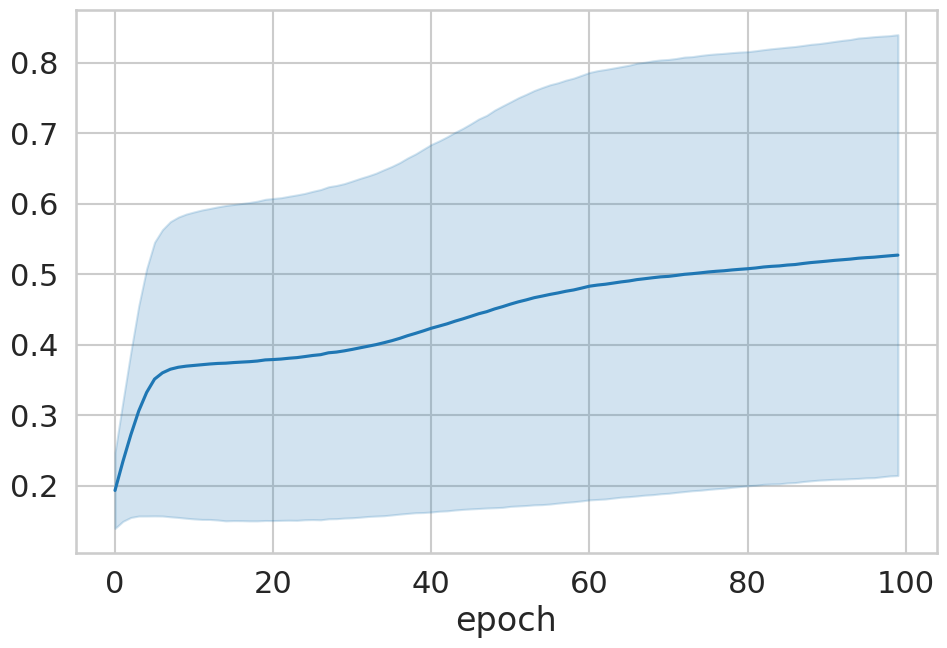}\par} &
{\centering \includegraphics[width=\linewidth]{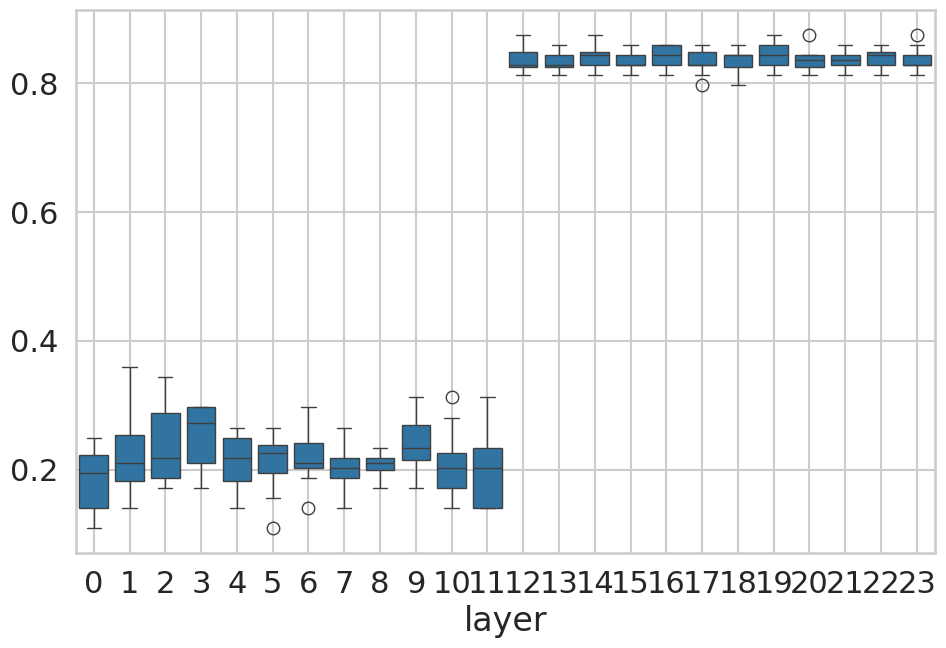}\par} \\[0.0 ex]

\multicolumn{4}{c}{\textbf{(e)}} \\[1.0ex]

{\centering \includegraphics[width=\linewidth]{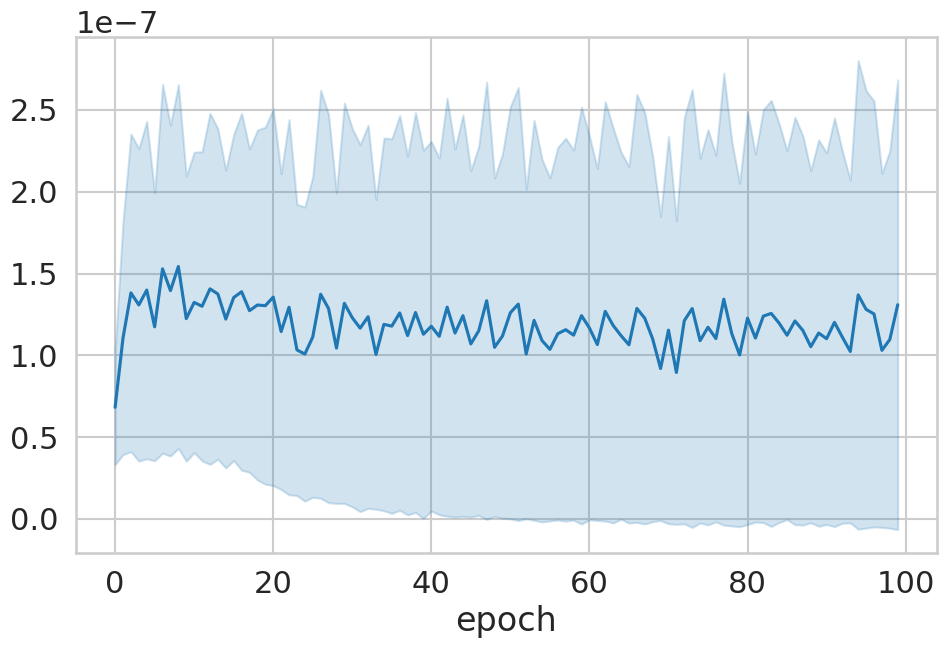}\par} &
{\centering \includegraphics[width=\linewidth]{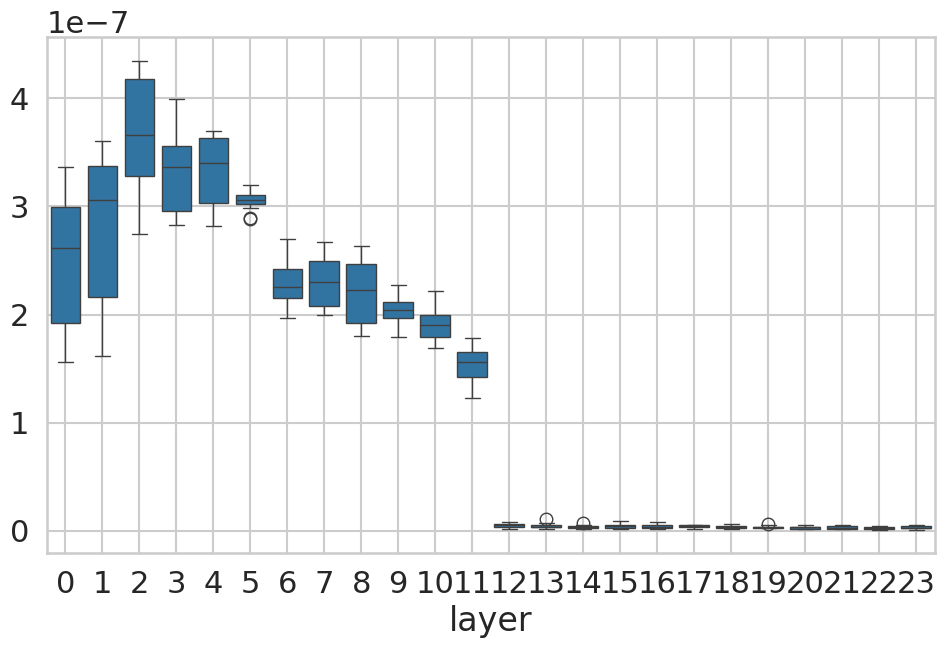}\par} &
{\centering \includegraphics[width=\linewidth]{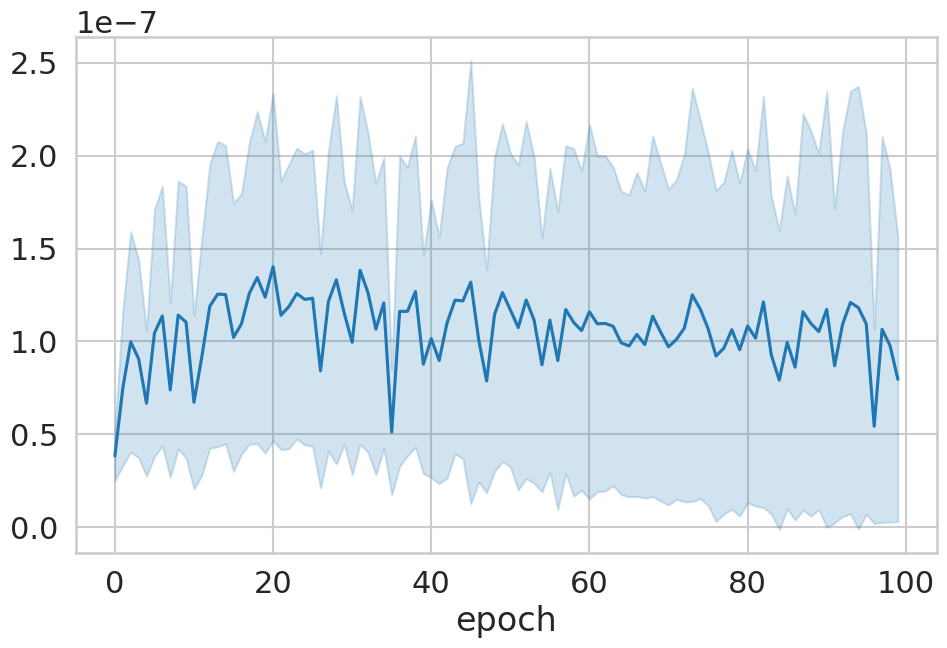}\par} &
{\centering \includegraphics[width=\linewidth]{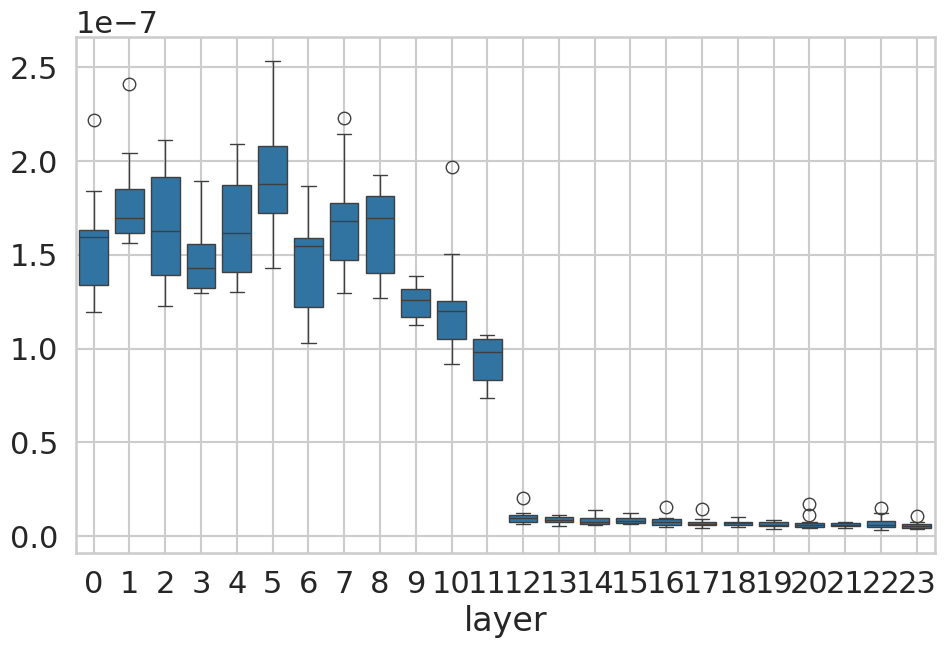}\par} \\[0.0 ex]

\multicolumn{4}{c}{\textbf{(f)}} \\

\end{tabular}

\caption{SVDA interpretability diagnostics across datasets.
Each row corresponds to one indicator:
(a) Spectral Entropy,
(b) Effective Rank,
(c) Angular Alignment,
(d) Selectivity Index,
(e) Spectral Sparsity,
(f) Perturbation Robustness.
For each dataset block (KITTI on the left, NYU-v2 on the right), the first column shows evolution across training epochs and the second column shows the corresponding per-layer distribution (box-plots).}
\label{fig:svda_grid_mde_full}
\end{figure*}

Temporal analysis of the interpretability indicators reported in \figurename~\ref{fig:svda_grid_mde_full} reveals features of the inner working of the Transformer architecture while training is evolving. Results show that spectral entropy and effective rank decrease steadily, signaling a reduction in the number of active latent directions. In parallel, spectral sparsity increases, indicating that uninformative spectral components are pruned.  Angular alignment and the selectivity index exhibit minimal variation across epochs, suggesting that directional coherence and token‑level focus are established early; perturbation robustness stabilizes quickly, implying that the model attains noise tolerance from the outset. These trends hold for both KITTI and NYU‑v2, and this is revealed only because SVDA was introduced in the Transformer architectures. The similarity of the results in different datasets reveals, in this way, that the core DPT architecture is robust. This conclusion is one of the benefits of using SVDA.

Furthermore, the per-layer distributions reveal a systematic depth-wise organization. Early layers retain higher entropy and rank, indicating broad spectral usage and diffuse attention, while deeper layers converge to low-rank, low-entropy profiles that emphasize a few dominant directions. Selectivity index increases with depth, showing that later layers route information more sharply across tokens, while early layers maintain distributed focus. Angular alignment is strongest in shallow layers and diminishes with depth. Perturbation robustness also increases in deeper layers, where attention becomes less sensitive to small input variations, thereby reinforcing stability near the decision stage. Again, all those results and conclusions would not be attainable in a totally opaque architecture (baseline), without introducing SVDA.



\section{Discussion}

The integration of SVDA into monocular depth estimation does not alter predictive behavior in a substantial way. Instead, it reframes the task from asking whether accuracy improves, to asking how the underlying architecture organizes its representational capacity. The results show that SVDA maintains parity with the baseline while providing structured descriptions of spectral usage, directional alignment, and robustness. This descriptive role is the central contribution expected by SVDA.

From this perspective, this paper showcases that SVDA serves as a diagnostic lens in monocular depth estimation. The observed reductions in entropy and rank, the rise of sparsity, and the layer-wise transitions reported are not prescriptive findings to be optimized toward, but descriptive patterns that reveal how the DPT backbone allocates attention across training and depth. Such patterns would remain opaque in conventional attention, where only dense, uninterpretable maps are available.

By shifting attention analysis into a spectral framework, SVDA aligns with broader efforts in explainable AI: it makes the organization of deep architectures observable without imposing external constraints or heuristic explanations. The importance of this shift is not in outperforming state-of-the-art metrics, but in providing a principled account of model dynamics that can be interrogated, compared across datasets, and eventually trusted in safety-critical applications.

In summary, the discussion does not concern accuracy gains, but transparency. SVDA describes, rather than prescribes, how attention behaves in monocular depth estimation, thus bridging its theoretical foundation with practical interpretability in vision Transformers.

\section{Conclusion}

This work applied SVD-Inspired Attention (SVDA) to the task of monocular depth estimation within a DPT framework. The results across KITTI and NYU-v2 confirm that SVDA preserves predictive accuracy while introducing a principled mechanism for examining the inner structure of attention. Unlike methods that seek architectural improvements or performance gains, SVDA provides a descriptive account of how attention spectra evolve across training and depth, revealing compression, sparsity, alignment, and robustness that remain hidden in standard formulations.

The contribution of this study is therefore twofold: first, it demonstrates that SVDA can be integrated into dense prediction architectures without loss of accuracy; second, it shows that the resulting interpretability indicators enable systematic inspection of representational dynamics. In doing so, SVDA transforms attention from an opaque mechanism into a structured and quantifiable process.

Future work will extend this descriptive perspective to other dense prediction tasks and to scenarios where interpretability is critical for trust and accountability. By grounding explanations in spectral analysis rather than heuristic visualization, SVDA offers a pathway toward attention models that can be not only accurate, but also meaningfully understood.

\bibliographystyle{unsrt}  
\bibliography{references}  


\end{document}